\documentclass[conference]{IEEEtran}
\IEEEoverridecommandlockouts

\usepackage{cite}
\usepackage{amsmath,amssymb,amsfonts}
\usepackage{algorithmic}
\usepackage{graphicx}
\usepackage{textcomp}
\usepackage{xcolor}

\usepackage{graphicx}  
\usepackage{multirow}  
\usepackage{tabularx}  
\usepackage{array}     
\usepackage{caption}   
\usepackage{amsmath}   
\usepackage{booktabs}  
\usepackage{float}     
\usepackage{graphicx}
\usepackage{amsmath} 
\usepackage{listings}
\usepackage{array}
\usepackage{subcaption}
\usepackage{soul}

\def\BibTeX{{\rm B\kern-.05em{\sc i\kern-.025em b}\kern-.08em
		T\kern-.1667em\lower.7ex\hbox{E}\kern-.125emX}}

\begin{document}

	\newcommand{\Surrogate}{NeuroLGP-SM}
	\newcommand{\Expensive}{NeuroLGP}

	\newcommand{\SurrogateCap}{NeuroLGP-SM}
	\newcommand{\ExpensiveCap}{NeuroLGP}
	
	\title{NeuroLGP-SM: Scalable Surrogate-Assisted
		Neuroevolution for Deep Neural Networks}

	\author{
		\IEEEauthorblockN{Fergal Stapleton$^*$\thanks{$^*$joint-first authorship}}
		\IEEEauthorblockA{{Naturally Inspired Computation Research Group,} \\
			{Department of Computer Science},
			{Hamilton Institute}\\
			Maynooth University, Ireland \\
			fergal.stapleton.2020@mumail.ie}
		\and
		\IEEEauthorblockN{Edgar Galv\'an$^*$}
		\IEEEauthorblockA{{Naturally Inspired Computation Research Group,} \\
			{Department of Computer Science},
			{Hamilton Institute}\\
			Maynooth University, Lero, Ireland \\
			edgar.galvan@mu.ie}
	}

	\maketitle
	
	\begin{abstract}
		
		
		%
		Evolutionary Algorithms (EAs) play a crucial role in the architectural configuration and training of Artificial Deep Neural Networks (DNNs), a process known as neuroevolution. However, neuroevolution is hindered by its inherent computational expense, requiring multiple generations, a large population, and numerous epochs. The most computationally intensive aspect lies in evaluating the fitness function of a single candidate solution. To address this challenge, we employ Surrogate-assisted EAs (SAEAs). While a few SAEAs approaches have been proposed in neuroevolution, none have been applied to truly large DNNs due to issues like intractable information usage. In this work, drawing inspiration from Genetic Programming semantics, we use phenotypic distance vectors, outputted from DNNs, alongside Kriging Partial Least Squares (KPLS), an approach that is effective in handling these large vectors, making them suitable for search. Our proposed approach, named Neuro-Linear Genetic Programming surrogate model (NeuroLGP-SM), efficiently and accurately estimates DNN fitness without the need for complete evaluations. NeuroLGP-SM demonstrates competitive or superior results compared to 12 other methods, including NeuroLGP without SM, convolutional neural networks, support vector machines, and autoencoders.  Additionally, it is worth noting that NeuroLGP-SM is 25\% more energy-efficient than its NeuroLGP counterpart. This efficiency advantage adds to the overall appeal of our proposed NeuroLGP-SM in optimising the configuration of large DNNs.


	\end{abstract}
	
	\begin{IEEEkeywords}
		Neuroevolution, Linear Genetic Programming, Surrogate-assisted Evolutionary Algorithms
	\end{IEEEkeywords}
	
	\section{Introduction}
	
	

	

	Evolutionary Algorithms (EAs)~\cite{EibenBook2003} have proven to be effective in both the crafting of architectures and hyperparameter optimisation of Deep Neural Networks (DDNs)~\cite{lecun2015deeplearning}. This application is commonly known as neuroevolution~\cite{Galvn2021NeuroevolutionID}, and have been applied to numerous problem domains, such as autonomous vehicles~\cite{galvan2023evolutionary,10.1145/3520304.3528984} and face recognition~\cite{10.1145/3520304.3528884,10177272}. The pursuit of optimal DNN architectures has led to various methodologies, including EAs~\cite{Galvn2021NeuroevolutionID}, reinforcement learning~\cite{DBLP:journals/corr/ZophL16}, etc. However, a significant challenge persists across these methods: the substantial computational resources required to identify high-performing networks.
	

	The rise of GPU-accelerated hardware has helped alleviate some of this computational cost, however, a significant proportion of research in DNNs is based on incremental improvements on DNN algorithms for benchmark problems, where there is a significant correlation between network complexity for incremental gains in terms of additional performance. In fact, when looking at very large models of hundreds of billions of parameters, it can cost millions of dollars for a single iteration~\cite{menghani2023efficient}. This energy consumption is further compounded when considering population-based neuroevolutionary techniques which require many networks to be trained and evaluated in order to find suitable architectures.
	
	One way to address this significant issue is with the use of surrogate-assisted evolutionary algorithms (SAEAs). SAEAs can be used to estimate the fitness of DNNs without the need to fully train each network. In particular, surrogate modelling strategies that employ Bayesian optimisation have shown much promise~\cite{menghani2023efficient}. However, a major challenge remains in how best to deal with the surrogate representation. For instance, using genotype information to build surrogates often requires complex encoding strategies~\cite{8744404}, and in some instances, constructing adequate distance metrics to compare different network topologies is not feasible~\cite{pheno_dist_kernel}. Using phenotypic information on the other hand has shown some promise~\cite{hagg2019prediction,pheno_dist_kernel}, but a challenge remains in scaling to deeper and more complex networks which inherently requires a high-dimensional representation~\cite{2023_GECCO_LBA_KPLS_FergalStapleton}.
	
	In this work, we analyse a novel population-based Neurovolutionary technique, referred to as NeuroLGP, and its surrogate model variant NeuroLGP-SM~\cite{stapleton2024ola}. Using a robust model management strategy, we use phenotypic distance vectors to estimate the performance of partially trained DNNs. These vectors are comparatively large for the optimisation problem at hand~\cite{2023_GECCO_LBA_KPLS_FergalStapleton} and, as such, we use an approach that is designed for handling high-dimensional data, known as Kriging Partial Least Squares (KPLS). This approach allows for a novel and scalable surrogate-assisted technique that is skilfully adept at handling neuroevolution of DNNs and to the the best of our knowledge, this method of surrogate-assisted neuroevolution has not been studied before.

	The aim of this study is to apply Surrogate-assisted Evolutionary Algorithms (SAEA) in neuroevolution, using Kriging Partial Least Squares (KPLS) on phenotypic distance vectors inspired by Genetic Programming Semantics. Our key contributions are outlined as follows, 1) \textbf{Efficient Fitness Estimation:} We demonstrate the accurate estimation of DNN fitness values without full evaluations through our proposed approach, NeuroLGP-SM, employing KPLS on phenotypic distance vectors. 2) \textbf{Performance Metrics:} We employ three well-defined metrics to assess the predictive capabilities of NeuroLGP-SM in terms of model fitness. These results align with the competitive or superior performance of our NeuroLGP-SM approach compared to 12 popular techniques, including convolutional neural networks, autoencoders, and support vector machines, across four challenging classification tasks. 3) \textbf{Energy Consumption Analysis:} We provide a reliable formula to gauge the energy consumption of our approaches, revealing that NeuroLGP-SM is 25\% more efficient compared to its counterpart that does not use surrogate models and 4) \textbf{Encoding for Analysis:} Through clever encoding, we allow easy access to analyse the internal structures of the architectures, enabling us to conduct an in-depth analysis of the networks discovered by our proposed approaches.

	\section{Related Work}
	\label{sec::related_work}

	\subsection{Neuroevolution using Surrogate Models}
	
	Santos et al.~\cite{santos2023neuroevolution} proposed a novel approach that makes use of semantics in neuroevolution. They do so by using Geometric Semantic Genetic Programming, in conjunction with a neuroevolutionary approach called Semantic Learning Machines (SLM)~\cite{gonccalves2015semantic}, a form of neuroevolution technique. Furthermore, Hagg et al.~\cite{hagg2019prediction}, made the connection between semantic distances in GP and phenotypic distances within the context of surrogate-assisted evolutionary algorithms for neuroevolution. Similarly, Stork et al.~\cite{pheno_dist_kernel} extended CGP to use a surrogate-assisted neuroevolution approach that makes use of phenotyic distance vectors. A limitation of Stork's work is the scalability of using Kriging on high-dimensional data. For instance, in the recent work by Stapleton and Galv{\'a}n, highlighted that traditional approaches such as the Kriging approach suffer with high-dimensionality~\cite{2023_GECCO_LBA_KPLS_FergalStapleton} and may not be suitable for DNN architectures.
	
	Freeze-Thaw Bayesian Optimisation
	(FBO), proposed by Swersky et al.~\cite{swersky2014freeze}, uses Bayesian optimisation to determine whether a particular neural network that has been partially trained should be fully evaluated. This approach is novel in that the system stops training (or freezing) of less promising networks, instead spending valuable resources on the most promising networks. Of note, is the fact that the FBO approach relies
	on the phenotypic behaviour of DNNs.
	It is important to note the majority of works rely on genotypic information when building surrogate models for neuroevolution~\cite{ 8744404,greenwood2022surrogate}. For reference, a major work in this regard is that of Sun et al.~\cite{8744404} referred to as the End-to-End Performance Predictor. This approach uses an offline surrogate model based on random forests to
	accelerate learning of CNN architectures. The CNN architecture is encoded such that it maps to a numerical decision variable, which is passed to the random forest-based surrogate-model. Not only does this approach not require large amounts of training time but, also
	alleviates the requirement of a smooth learning curve~\cite{swersky2014freeze}.

	
	\section{Background}
	\label{sec::background}

	\subsection{Surrogate assisted models: Kriging and Kriging Partial Least Squares }
	\label{sec::kriging}
	
	Surrogate modelling has many use cases but in the context of this work, the aim is to effectively estimate the fitness values for candidate solutions while simultaneously reducing the run time of the evolutionary process. To this end, not only must the surrogate model be well-posed, but also, the evolutionary process must interact with a robust surrogate model management strategy~\cite{jin2011surrogate}. The surrogate model differs from the parent model that instead of training directly on the data the surrogate model is trained on the design space, where the aim is to identify and further explore regions of this design space that will produce preferable parameters. As such, interpolation-based approaches may be used to simulate the unknown regions of the parameter space, such as Gaussian processes, commonly referred to as Kriging.  Another benefit to the Kriging approach is that it allows for estimates of the uncertainty of predictions. 
	
	
	Kriging is an interpolation-based technique that assumes spatial correlation exists between known data points, based on the distance, and variation between these points. We aim to use observations $\{y(x_1), y(x_2),...y(x_n) \}$ to help estimate an unknown function value $\hat{y}(x^{*})$ for the unknown data point $x^{*}$, where $n$ is the number of individuals in the training data. A kernel function $K(\cdot)$ is used to express the spatial correlation between two samples $x_i$ and $x^{'}_{i}$ as shown in Eq.~\ref{eqn::kernel_dist},
	
	\begin{equation}
	k(x, x') = \prod_{i=1}^m exp(- \theta (\mathcal{D}(x_i, x^{'}_i))) 
	\label{eqn::kernel_dist}
	\end{equation}
	
	\noindent where the $\theta$ parameter controls the rate at which the correlation decays to zero between $x_i$ and $x^{'}_{i}$ and $\mathcal{D}(x_i, x^{'}_i) = (x_i - x^{'}_i)^2$. The $\theta$ parameter is determined using the Maximum Likelihood Estimator (MLE). A significant drawback, however, is that for large dimensions $m$, the computational cost increases dramatically since the MLE algorithm calculates the inverse of the correlation matrix multiple times.


	Kriging Partial Least Squares (KPLS) helps to alleviate this limitation by reducing the number of parameters calculated~\cite{bouhlel2016improving}. It does so by using partial least squares which project the high-dimensional data into a lower dimension using principal components. Eq.~\ref{eqn::kernel_dist2} details the KPLS kernel,

	\begin{equation}
	K(x, x') = \prod_{k=1}^{h} \prod_{i=1}^{m} exp(- \theta_k (w^{(k)}_{i} x_i - w^{(k)}_{i} x'_i)^2)
	\label{eqn::kernel_dist2}
	\end{equation}
	
	\noindent where $w$ are rotated principal directions which maximise the covariance and are a measure of how important each principal component is. Typically, the number of principal components $h$ is much less than the number of dimensions $m$ and, as such, allows for the substantial improvement in computational cost associated with the KPLS approach. See~\cite{bouhlel2016improving} for details.
	
	\subsection{Phenotypic Distance}
	
	
	The inspiration for using phenotypic distance is by works such as Stork et al.~\cite{pheno_dist_kernel} and Galv{\'a}n et al.~\cite{9308386,galvan2022semantics,DBLP:conf/gecco/GalvanS19,stapleton2021semantic}, whose previous work in GP centred on semantics, in particular, the use of semantic distance metrics. In traditional GP, semantics can be understood as the behaviour of program given a finite set of inputs. A definition by Pawlak et al.~\cite{6808504} states

	\textbf{Def 1.} \emph{The semantics $s(p)$ of a program $p$ is the vector of
		values from the output set $O$ obtained by computing $p$ on all inputs from the input set $I$}. This is  expressed in Eq.~\ref{eqn::def1},
	
	\begin{equation}
	s(p) = [p(in_1), p(in_2), ... , p (in_l)]
	\label{eqn::def1}
	\end{equation}
	
	\noindent where $l = |I|$ is the size of the input set and where the inputs $in$ from can be some subset of the dataset. In terms of neuroevolution, we can define a solution sample $x$ as having the semantic or phenotypic behaviour of the $i^{th}$ program such that $x_i = s(p_i)$, where the semantics $s(p)$ is the vector of values from the output. From Eq.~\ref{eqn::kernel_dist} we can then define our phenotypic distance $\mathcal{D}$ as,
	
	\begin{equation}
	\mathcal{D}(x_i, x_j) = \mathcal{D}(s(p_i), s(p_j)))
	\label{eqn::kernel_dist3}
	\end{equation}
	
	
	As such the distance metric is dependent on the outputs of each network, where the $x_i$ is a flattened vector containing the outputs from the final layer for all data instances. Thes phenotypic distance vector length, as such, is the number of images of the test dataset times the number of classes. Notably, this approach can be extended to any deep learning model architecture that uses a vectorised output (i.e., transformers), however, the limitations and scalability of other deep learning models with this method are yet to be investigated.
	
	\section{Methodology}
	\label{sec::methodology}
	
	\subsection{NeuroLGP}
	\label{sub::neurolgp}



	The original Linear Genetic Programming (LGP) encoding is based on the concept of using registers, which are units computer memory storage for manipulating data while executing instructions, in a low-level programming context. The content of these registers are altered using instruction operations. There are three main components to instruction; an \textit{operand} which performs a specific function on one or more \textit{registers} which store the result in a \textit{destination register}. In the case of 2-register instruction encoding the operand operates on a single instruction and for a 3-register instruction encoding operates on two instructions. The left of Fig.~\ref{fig:lgp_neuro_example} highlights an example of LGP written in C code, where {\fontfamily{pcr}\selectfont{r[i]}} denotes the $i^{th}$ register. This example contains both effective and non-effective lines of code, where the non-effective lines of code are commented out and subsequently not compiled. Each line of code is executed imperatively. The register {\fontfamily{pcr}\selectfont{r[0]}} is a specially designated register for the final output of the program. With the NeuroLGP approach, instead of using registers to store small amounts of data, we instead use the idea of registers as pointers to much larger amounts of data. As such, the registers instead control the flow of the initial and intermediary data from each outputted layer of our evolvable DNN. The right of Fig.~\ref{fig:lgp_neuro_example}, demonstrates how the expected representation would look, where the general form of this code is similar to the figure on the left.

\begin{figure}
	
\noindent\begin{minipage}{.235\textwidth} 
\begin{lstlisting}[basicstyle=\scriptsize,,frame=tlrb]
void gp(r)
double r[8];
{ ...
   r[0] = r[5] + 71;
   // r[7] = r[0] - 59;
   if (r[1] > 0)
   if (r[5] > 2)
   r[4] = r[2] * r[1];
   // r[2] = r[5] + r[4];
   r[6] = r[4] * 13;
   r[0] = r[6];
}
\end{lstlisting}
		
\end{minipage}\hfill
\begin{minipage}{.235\textwidth}

		
\begin{lstlisting}[basicstyle=\scriptsize,,frame=tlrb]
def neuro_gp(...)
		
{  ...
   r[0] := Conv(r[1])
   // r[4] := MaxPool(r[3])
   // r[7] := Conv(r[4])
   r[5] := MaxPool(r[2])
   r[6] := Conv(r[5])
   r[3] := AvgPool(r[6])
   // r[5] := MaxPool(r[8])
   r[0] := Dense(r[3])  
}
		
\end{lstlisting}
		
	\end{minipage}\hfill
	\caption{\textit{left:} LGP example in C language. Based on example from~\cite{brameier2007linear}. \textit{right:}  NeuroLGP psuedocode for python.
	}
	\label{fig:lgp_neuro_example}
	\end{figure}

\begin{figure*}[tb]
		\centering
		
		\includegraphics[width=0.8\linewidth]{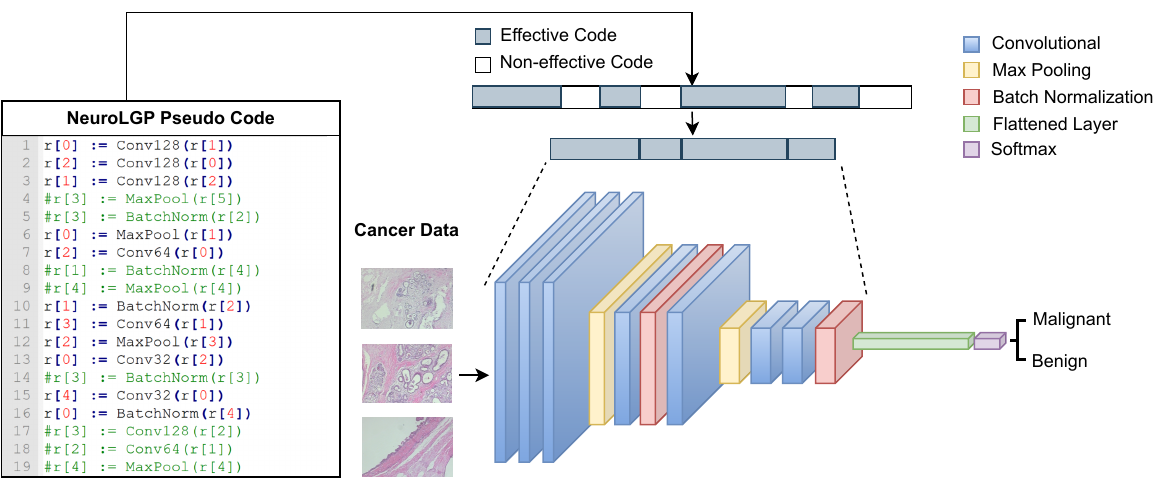}

		\caption{Diagram of the NeuroLGP genotype-to-phenotype mapping. The pseudocode for the set of instructions (left-hand side) can be represented as the genotype with effective and non-effective code (top) and produces the resulting phenotype (bottom right-hand side) as a specific neural network architecture. Note that the non-effective coding is not present in the phenotype.}
		
		\label{fig:phenotype}
	\end{figure*}

	Traditionally, in LGP, registers can be of different types (input registers, calculation registers and constant registers). In this work, each register is initialised with the inputs of various layers within the neural network and so act as just input registers (though it may be useful to have special registers specifically for handling skip connections and multiple branches). While this may seem wasteful in terms of memory, multiple registers are only used in the genotypic representation and when calculating the fitness, the number of effective registers can be reduced as part of a repair process. Likewise, the non-effective code can also be removed at the same time. As such, the phenotypic representation of the network will not only be much simpler than the genotypic representation, but also much smaller. Fig.\ref{fig:phenotype}
	demonstrates the genotype-to-phenotype mapping.

	\subsection{NeuroLGP with Surrogate Model (NeuroLGP-SM)}
	\label{sub::neurolgp-sm}
	
\begin{figure}[tb]
		\centering
		
		\includegraphics[width=0.9\linewidth]{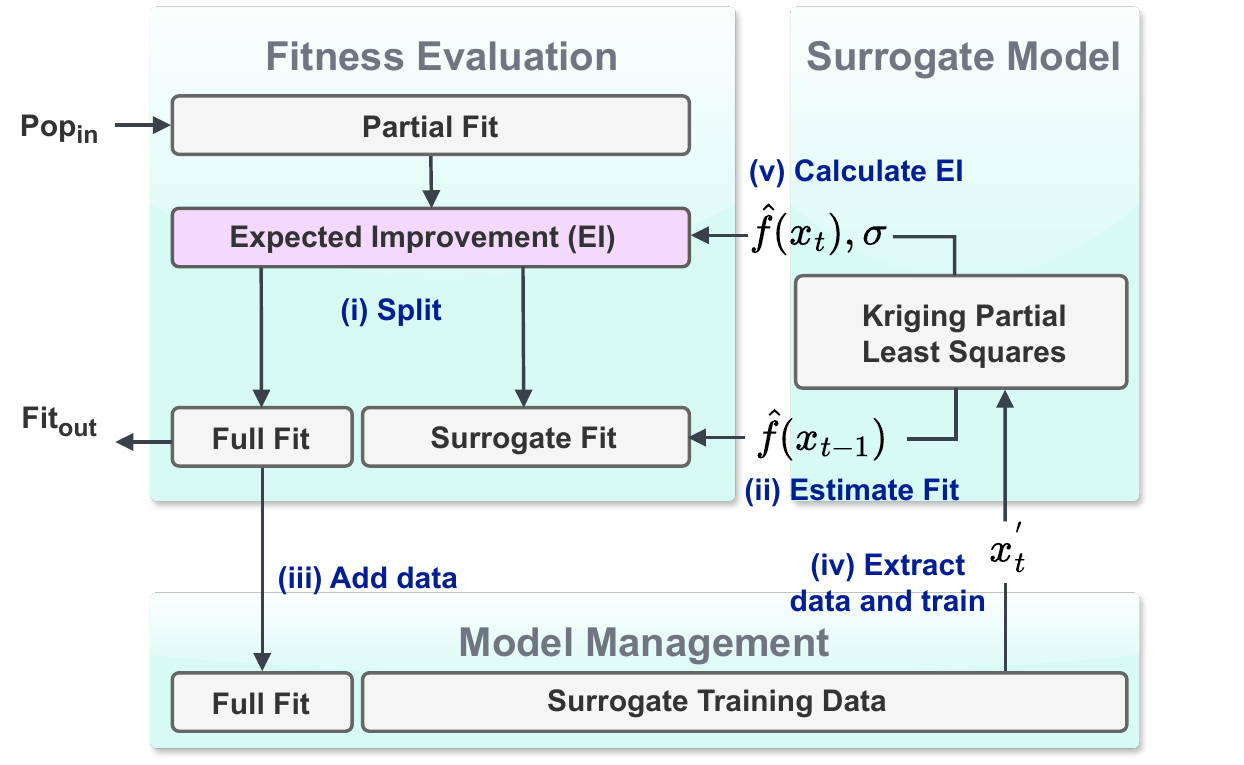}

		\caption{Model management strategy in terms of fitness evaluation, model management and training the surrogate model.}
		
		\label{fig:surrmodel}
	\end{figure}

	To identify individuals requiring full evaluation, we use the Expected Improvement (EI) criteria~\cite{jones1998efficient}. EI guides the selection of candidate solutions for evaluation by estimating the improvement over the current best solution. This enables us to prioritise solutions from areas in the search space expected to exhibit the most significant improvement. The calculation of EI is shown in Eq.~\ref{eqn::ei},
	
	\begin{equation}
	\begin{aligned}
	\text{EI} &= \left\{
	\begin{aligned}
	&( \hat{f}(x) - f(x^{*}) ) \Phi(Z) + \sigma(x) \phi(Z) && \quad  \text{if}\ \sigma(x) > 0  \\
	&0 && \quad \text{if}\ \sigma(x) = 0
	\end{aligned}
	\right. \\
	Z &= \frac{ \hat{f}(x) - f(x^{*}) }{\sigma(x) }
	\end{aligned}
	\label{eqn::ei}
	\end{equation}

	\noindent where $\Hat{f}(x)$ is the model's predicted performance of the surrogate for the phenotypic distance vector $x$, where $f(x^{*})$ is the best-known value of the objective function so far (in this case maximum) and $\Phi$ and $\phi$ are the cumulative distribution function (CDF) and probability density function (PDF) of the standard normal distribution, respectively. Fig.~\ref{fig:surrmodel} summarises the surrogate model management strategy. The annotations are: \textit{(i) Split:} first, the population is split with a 40/60 split for individuals to be fully evaluated \textit{vs.} the partially trained individuals, respectively, \textit{(ii) Estimate fitness:} the fitness is estimated using the KPLS approach as informed by the previous generation, \textit{(iii) Add data:} the phenotypic distance vector for the fully evaluated portion of the population is added to the surrogate training data (note: the phenotypic distance vector is taken before the full fitness evaluation), \textit{(iv) Extract data and train:} the phenotypic distance vectors are collectively used to train the KPLS approach as detailed in Section \ref{sec::kriging}, and \textit{(v) Calculate EI:} the EI criteria is calculated for the incoming population of individuals.

	\section{Experimental setup}
	\label{sec::exp}
	
	
	The Breast Cancer Histopathological Image Classification (BreakHis)~\cite{spanhol2016breast} is a binary classification dataset consisting of microscopic images containing 2,480 benign and 5,429 malignant tumours, where images were obtained as four different magnifications ($\times$40, $\times$100, $\times$200 and $\times$400) and are split into four individual datasets for this work. Each image consists of 64x64 pixels (down-scaled resolution from 700x460 pixels) and 3-channel RGB with 8-bit depth in each channel. The split for training, validation, test and test2 is approximately (63.5\%, 12.5\%, 12.5\%, 12.5\%), where the first three splits conform to 
	a ratio of (70:15:15), conforming to the general training/test split as seen in other approaches (Table ~\ref{tab::results}). The first three splits are used purely for training and evaluating networks and the last split is retained for an unbiased analysis of the evolutionary process. The dataset is imbalanced and the synthetic minority oversampling technique~\cite{chawla2002smote} is used to up-sample the minority class.
	
	
	
	
	
	
	
	
	We employ two approaches to investigate the validity and quality of our proposed surrogate-assisted neuroevolution model. These are: \textit{(i)} The \textit{\Expensive{}} approach, which uses a novel encoding to evolve the structure of our networks, training to the full number of epochs as outlined in Section \ref{sub::neurolgp}, and \textit{(ii)} The \textit{\Surrogate{}} approach where we integrate the surrogate modelling strategy, as outlined in Section~\ref{sub::neurolgp-sm}, into the \textit{\Expensive{}} approach. Experiments were conducted using Kay supercomputer provided by the Irish Centre for High-End Computing (ICHEC). Experiments were run in parallel, with each run assigned to a single Nvidia Tesla V100 GPU with 16GB Ram. Additionally, each run has access to a 20-core
	2.4 GHz Intel Xeon Gold 6148 (Skylake) CPU processor which is used during training of the surrogate model portion of the \Surrogate{} approach. Overall, the \Expensive{} approach took $\sim$28 GPU days and the \Surrogate{} approach
	$\sim$21 GPU days, for 8 runs across the 4 datasets.
	

	\section{Results}
	\label{sec::results}
	
	\subsection{Comparison of Models}
	
	\newcommand{\maxRanXforty}{0.889}
	\newcommand{\maxRanXhundred}{0.869}
	\newcommand{\maxRanXtwohundred}{0.946}
	\newcommand{\maxRanXfourhundred}{0.914}
	
	\newcommand{\maxSurXforty}{0.913}
	\newcommand{\maxSurXhundred}{0.903}
	\newcommand{\maxSurXtwohundred}{0.970}
	\newcommand{\maxSurXfourhundred}{0.925}
	
	\newcommand{\maxExpXforty}{0.930}
	\newcommand{\maxExpXhundred}{0.916}
	\newcommand{\maxExpXtwohundred}{0.960}
	\newcommand{\maxExpXfourhundred}{0.925}
	
	
	
	Table~\ref{tab::results} summarises approaches that have used the BreakHis data set from the last few years, including our proposed approaches. Most of these works have been taken from a comprehensive 2020 review paper from Benhammou et al.~\cite{benhammou2020breakhis}. For fair comparison, we have selected works that also do not use transfer learning as these represent manually crafted networks that have zero pre-training. Furthermore, many of these approaches use specialised feature extraction and pre-processing steps for histopathological data. As such, our goal is not necessarily to outperform previous state-of-the-art works, but rather to show that our neuroevolutionary technique can achieve similar results to hand-crafted architectures, even without specialised knowledge of the problem domain.

	Looking at the last four columns from the right, we list the accuracies for the various magnifications. The last four rows correspond to our proposed approaches: NeuroLGP and its surrogate-assisted variant (NeuroLGP-SM). We can see that our results are in good accordance with the other works, in some cases outperforming other approaches. Notably, the results yielded by our two approaches on the $\times$200 magnification are very competitive in terms of accuracy. 
	
	If we turn our attention to specifically comparing the \Surrogate{} and \Expensive{} models, we can see that each has very similar performances. For instance, the mean values (third and fourth last lines from bottom), the \Surrogate{} and \Expensive{} models are typically within 0.2$\%$ of each other and for the best (last two lines) the \Surrogate{} and \Expensive{} models are within 1-2$\%$.

	\begin{table*}[h]
		\centering
		\caption{Accuracy results for approaches using BreakHis dataset ($\times40$, $\times100$, $\times200$ and $\times400$). Results from this work are highlighted in boldface and are presented in the last four rows. WSI: wholes slide image, CNN: convolutional neural network, SVM: support vector machine, AE: auto-encoder, DBN: deep-belief network.}
		\resizebox{0.98\linewidth}{!}{
			\label{tab::results}
			\begin{tabular}{lcccccccc}
				\hline
				Work & Preprocessing & Patch/slide & Model & Training/Test & $\times$40 & $\times$100 & $\times$200 & $\times$400 \\
				\hline
				Gupta~\cite{gupta2017integrated} & None & WSI & Ensemble & 70\% / 30\% & 88.9 & 88.9 & 88.9 & 88.9 \\
				Sharma~\cite{sharma2017classification} & Mixed & WSI & Ensemble & Not specified & 81.7$\pm$2.8 & 81.2$\pm$2.7 & 80.7$\pm$3.4 & 81.5$\pm$3.1 \\
				Nahid~\cite{nahid2018histopathological} & None & WSI & CNN & Not specified & 94.4 & 95.93 & 97.19 & 96 \\
				Nahid~\cite{nahid2018histopathological2} & K-Mean Clustering & WSI & CNN & Not specified & 85 & 90 & 90 & 90 \\
				Karthiga~\cite{karthiga2018automated} & K-Mean Clustering & WSI & SVM & Not specified & 93.3 & 93.3 & 93.3 & 93.3 \\
				Pratiher~\cite{pratiher2019diving} & Gray Scale, Data Aug & WSI & AE & Not specified & 96.8 & 98.1 & 98.2 & 97.6 \\
				Badejo~\cite{badejo2018medical} & None & WSI & SVM & 70\% / 30\% & 91.1 & 90.7 & 86.2 & 84.3 \\
				Nahid~\cite{nahid2018histopathological3} & Contrast Enhancement & WSI & DBN & 70\% / 30\% & 88.7 & 89.06 & 88.84 & 87.67 \\
				Das~\cite{das2018multiple} & Resize (370x230) & Patches (224x224) & CNN & 80\% / 20\% & 89.52 & 85.3 & 88.6 & 88.4 \\
				Kumar~\cite{kumar2018breast} & Stain Normalisation & Patches (64x64, 32x32) & CNN & 70\% / 30\% & 82$\pm$2.8 & 86.2$\pm$4.6 & 84.6$\pm$3.0 & 84$\pm$4.0 \\
				
				Aswathy~\cite{aswathy2021svm} & Mixed & WSI & SVM & 90\% / 10\% & 89.1 & 89.1 & 89.1 & 89.1 \\
				
				\textbf{NeuroLGP (Mean)} & Resize (64x64), Data Aug & WSI & CNN & See Section \ref{sec::exp} & 89.7$\pm$2.1 & 0.872$\pm$3.0 & 92.6$\pm$2.1 & 90.8$\pm$1.0 \\
				\textbf{NeuroLGP-SM (Mean)} & Resize (64x64), Data Aug & WSI & CNN & See Section \ref{sec::exp} & 89.8$\pm$1.4 & 0.873$\pm$1.8 & 92.8$\pm$2.4 & 91.0$\pm$0.7 \\
				\textbf{NeuroLGP (Best)} & Resize (64x64), Data Aug & WSI & CNN & See Section \ref{sec::exp} & 93 & 91.6 & 96 & 92.5 \\
				\textbf{NeuroLGP-SM (Best)} & Resize (64x64), Data Aug & WSI & CNN & See Section \ref{sec::exp} & 91.3 & 90.3 & 97 & 92.5 \\
				\hline
			\end{tabular} 
		}
	\end{table*}
	
	\subsection{Analysis of the \SurrogateCap{} Model}
	
	\newcommand{\mseXforty}{0.0037}
	\newcommand{\mseXhundred}{0.0017}
	\newcommand{\mseXtwohundred}{0.0014}
	\newcommand{\mseXfourhundred}{0.0009}
	
	\newcommand{\kendallXforty}{0.6019}
	\newcommand{\kendallXhundred}{0.6791}
	\newcommand{\kendallXtwohundred}{0.6225}
	\newcommand{\kendallXfourhundred}{0.5647}
	
	\newcommand{\codXforty}{0.5026}
	\newcommand{\codXhundred}{0.6665}
	\newcommand{\codXtwohundred}{0.7079}
	\newcommand{\codXfourhundred}{0.7786}
	
	\newcommand{\timeSurXforty}{15.9}
	\newcommand{\timeSurXhundred}{16.6}
	\newcommand{\timeSurXtwohundred}{16.8}
	\newcommand{\timeSurXfourhundred}{15.3}
	
	\newcommand{\timeSurStdXforty}{0.7}
	\newcommand{\timeSurStdXhundred}{0.7}
	\newcommand{\timeSurStdXtwohundred}{0.8}
	\newcommand{\timeSurStdXfourhundred}{0.6}
	
	\newcommand{\timeExpXforty}{21.3}
	\newcommand{\timeExpXhundred}{22.7}
	\newcommand{\timeExpXtwohundred}{22.4}
	\newcommand{\timeExpXfourhundred}{20.0}
	
	\newcommand{\timeExpStdXforty}{0.2}
	\newcommand{\timeExpStdXhundred}{0.1}
	\newcommand{\timeExpStdXtwohundred}{0.1}
	\newcommand{\timeExpStdXfourhundred}{0.1}
	
	\newcommand{\timePercXforty}{25.3}
	\newcommand{\timePercXhundred}{26.7}
	\newcommand{\timePercXtwohundred}{24.9}
	\newcommand{\timePercXfourhundred}{23.8}
	
	Three metrics are used to determine how well the surrogate model performs in predicting
	the fitness of our partially trained models: \textit{(i)} the Mean Squared Error (MSE) gives a measure of how accurate our model is in terms of predicting fitness where values close to 0 are preferable, \textit{(ii)} Kendall's Tau is used to measure the correlation between the predicted fitness and the actual fitness~\cite{sun2019evolving}, where values close to 0 indicates no correlation, -1 a perfect negative correlation, +1 a perfect positive correlation, and \textit{(iii)} the R$^2$ score is a measure of how close the predicted value is to its true value and ranges from -$\infty$ to 1 and explains how much variability is in the prediction model, where values closer to 1 are preferable.

	
	Table~\ref{tab:surrogate_effectiveness} summarises the effectiveness of the \Surrogate{} model for each of the datasets. The low MSE values, as seen in the first row, show a relatively low error between the predicted and actual fitness across the four dataset. The Kendall's Tau values range from \kendallXfourhundred{} to \kendallXhundred{}, indicating a strong positive correlation between the actual and predicted fitness. We can see that for $\times$40 we have R$^2$ of \codXforty{} indicating a moderate level of fit being captured by the surrogate model while for $\times$100, $\times$200 and $\times$400, we have R$^2$ values of \codXhundred{}, \codXtwohundred{} and \codXfourhundred{}. There is also a general trend showing the R$^2$ score increasing with higher magnifications. This may be a result of more noise/artefacts present in the lower magnification images, making it more difficult for the surrogate model to predict based on the network outputs but further studies would be required to confirm this. Overall, the quality of fit and performance of the surrogate models based on the three metrics is very strong for each dataset.
	
\begin{figure}[tb]
		\centering
		
		\includegraphics[width=0.85\linewidth]{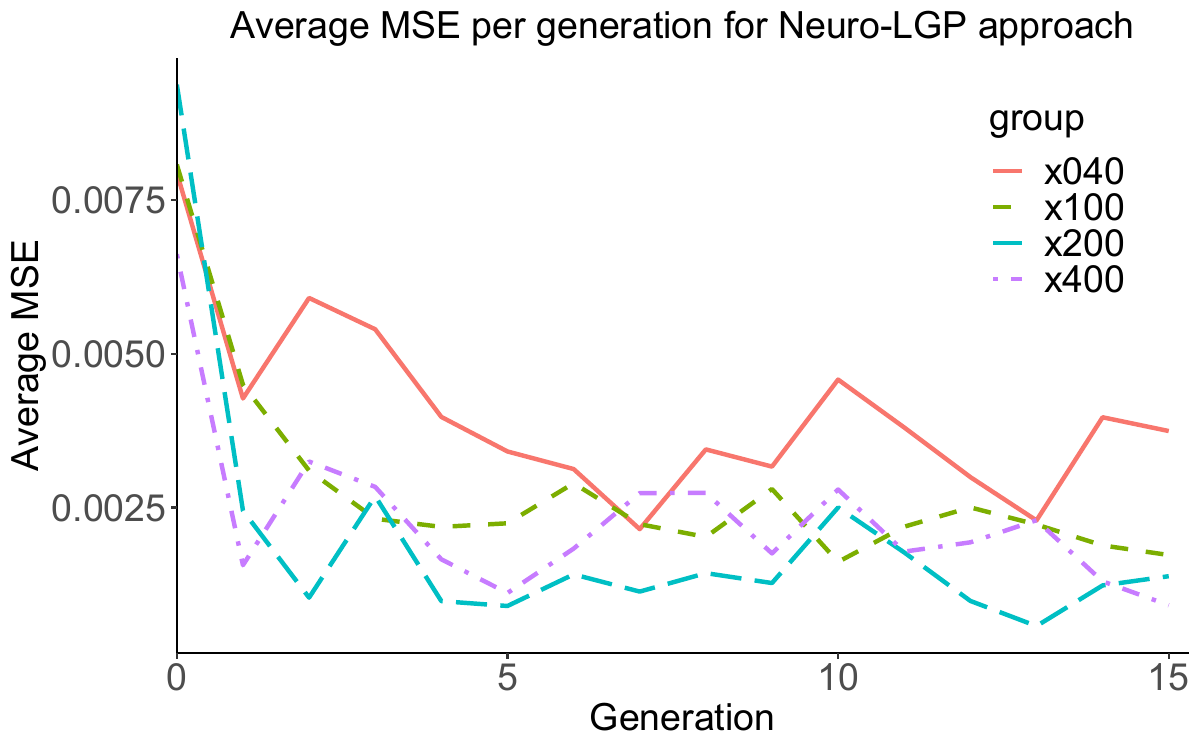}
		\caption{Avg. MSE between predicted \textit{vs.} actual fitness over 15 generations for the \Surrogate{} approach. The relative stability shows the robustness of the \Surrogate{} approach.}
		
		\label{fig::surrogate_mse}
	\end{figure}

	Next, we plot the average MSE per generation to get an idea of how the predictive capability of the \Surrogate{} approach changes over time as seen in Fig.~\ref{fig::surrogate_mse}. MSE values that either decrease or remain stable are preferable, as increasing MSE values would indicate our surrogate model is losing its predictive capability as new individuals are introduced. We can see that after an initial drop in average MSE from the first couple of generations (x-axis), while there are some fluctuations in the average MSE, in general values remain relatively stable, demonstrating the robustness of our \Surrogate{}.
	
	\addtolength{\tabcolsep}{+2pt}  
	\begin{table}[h]
		\centering
		\caption{Average MSE, Kendall's Tau and R$^2$.}
		\begin{tabular}{lcccc}
			\hline
			\multirow{2}{*}{Metric} & \multicolumn{4}{c}{Magnification Size} \\
			\cline{2-5}
			& $\times$40 & $\times$100 & $\times$200 & $\times$400 \\
			\hline
			MSE & \mseXforty & \mseXhundred & \mseXtwohundred & \mseXfourhundred \\
			Kendall's Tau &\kendallXforty & \kendallXhundred & \kendallXtwohundred & \kendallXfourhundred \\
			R$^2$ & \codXforty & \codXhundred & \codXtwohundred & \codXfourhundred \\
			\hline
		\end{tabular}
		\label{tab:surrogate_effectiveness}
	\end{table}
	
\begin{figure*}[htbp!]
		\begin{subfigure}[b]{0.245\textwidth}
			\includegraphics[width=\textwidth]{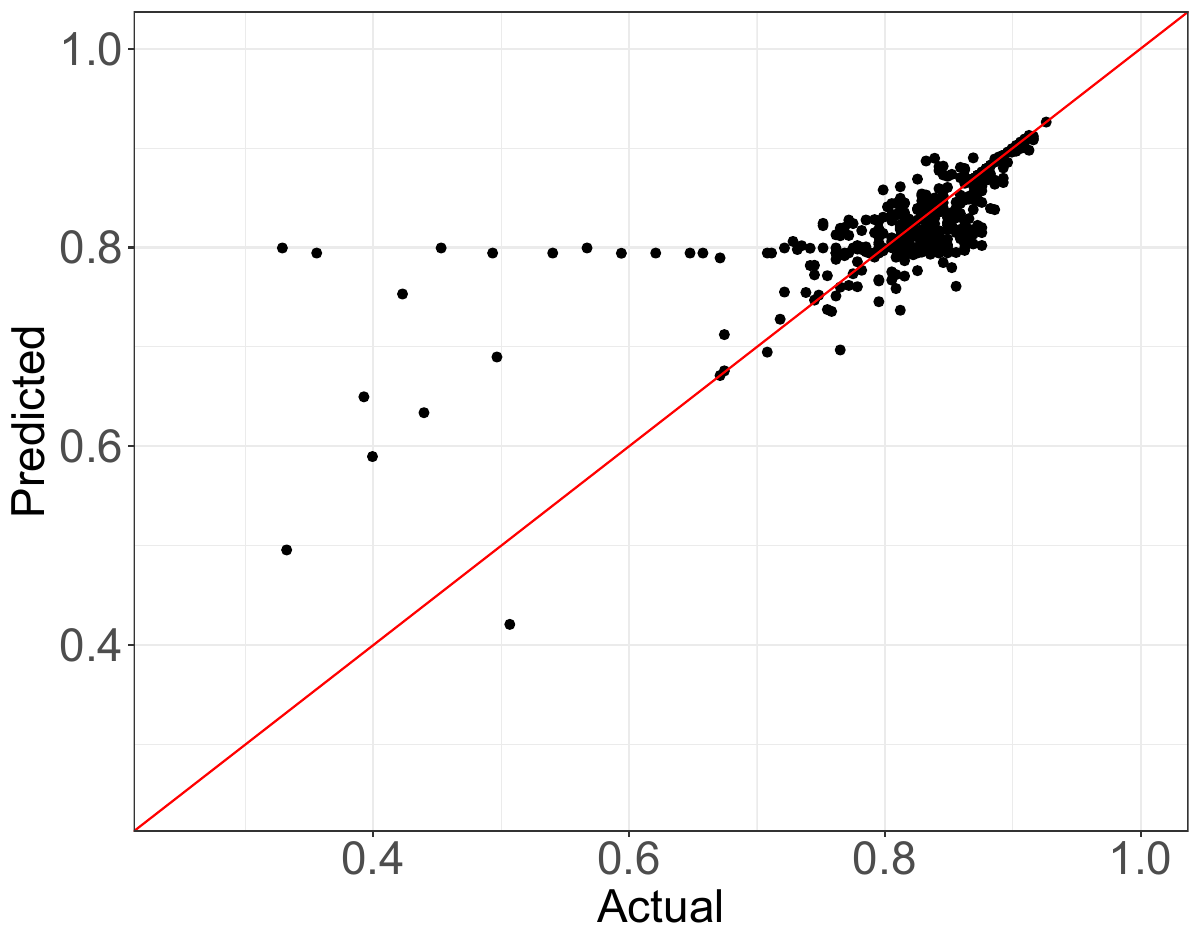}
			\caption{$\times$40}
			\label{fig:cod40}
		\end{subfigure}%
		\hfill
		\begin{subfigure}[b]{0.245\textwidth}
			\includegraphics[width=\textwidth]{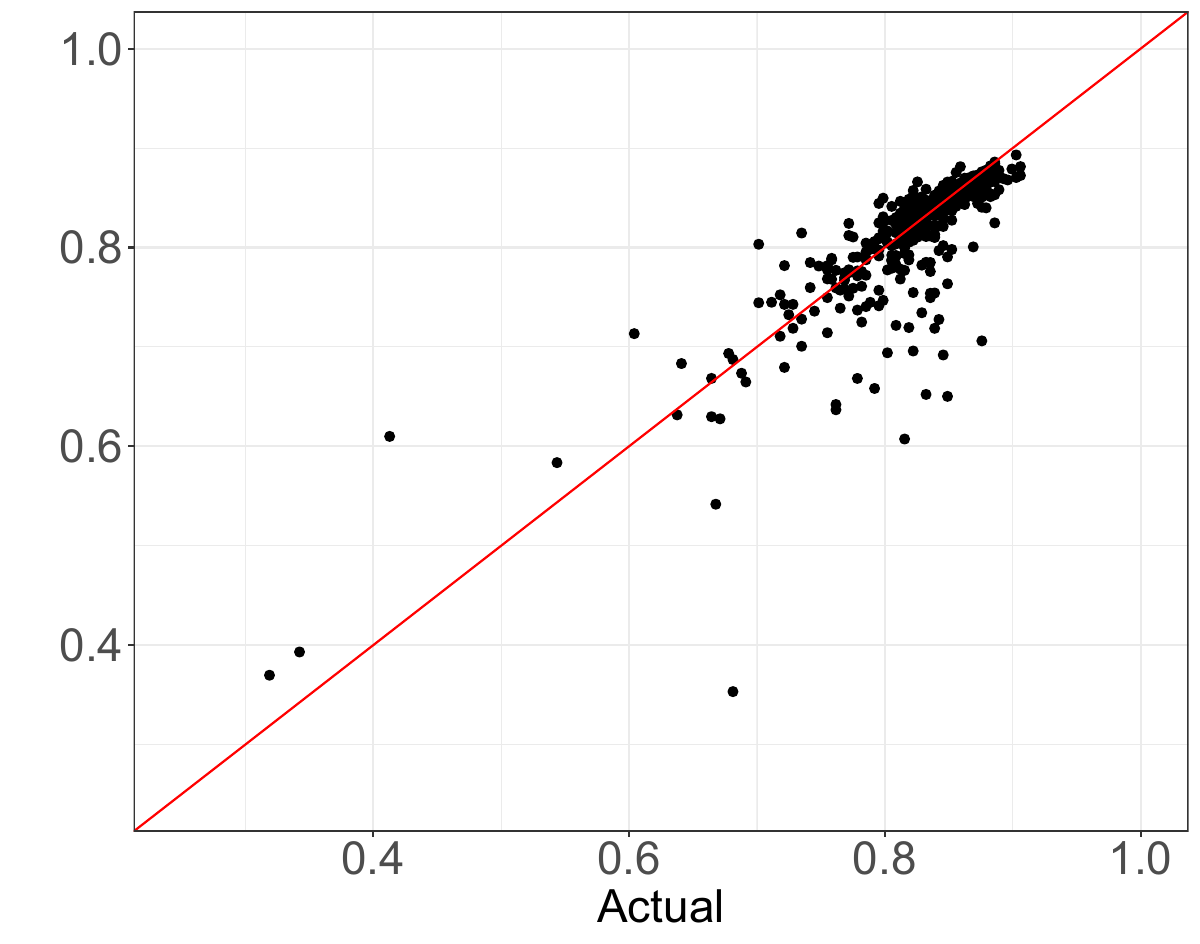}
			\caption{$\times$100}
			\label{fig:cod100}
		\end{subfigure}%
		\hfill
		\begin{subfigure}[b]{0.245\textwidth}
			\includegraphics[width=\textwidth]{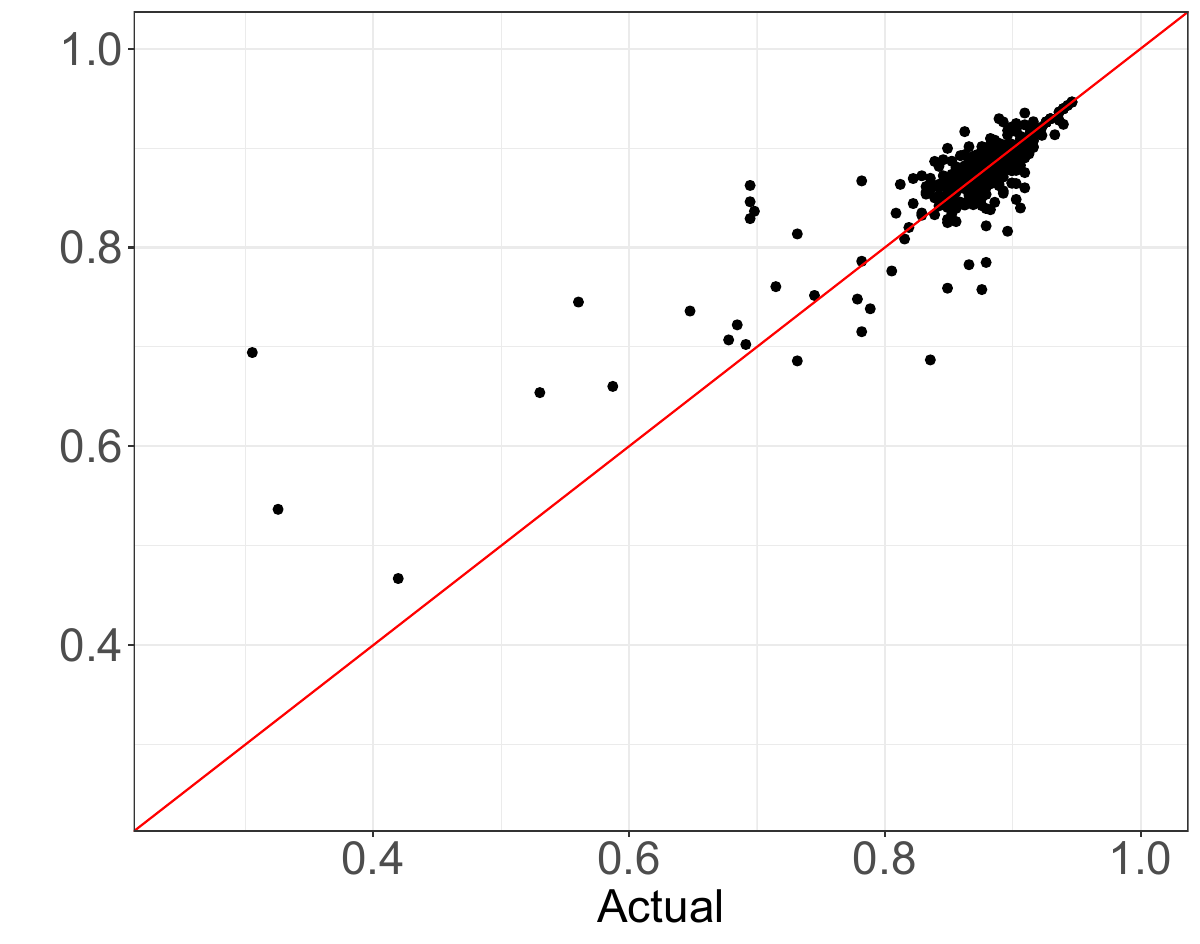}
			\caption{$\times$200}
			\label{fig:cod200}
		\end{subfigure}%
		\hfill
		\begin{subfigure}[b]{0.245\textwidth}
			\includegraphics[width=\textwidth]{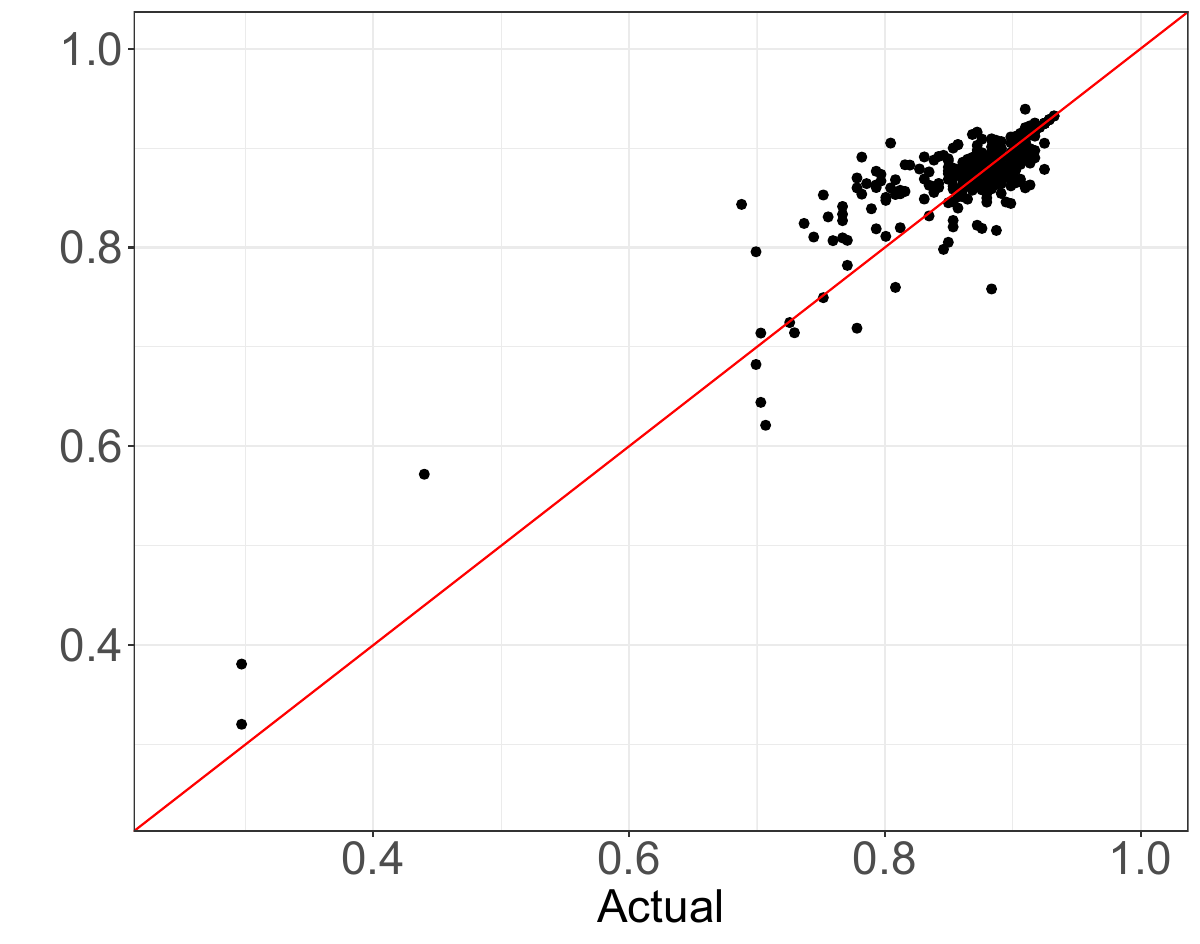}
			\caption{$\times$400}
			\label{fig:cod400}
		\end{subfigure}%
		\caption{ Predicted \textit{vs.} actual accuracies (black points), for $\times$40, $\times$100, $\times$200, and $\times$400 datasets, shown in (a) -- (d), respectively, across 8 independent runs for the surrogate-assisted approach (NeuroLGP-SM). The red line denotes where the accuracy for both predicted and actual are the same, where points closer to this line are preferential.}
		\label{fig:2x2cod}
	\end{figure*}

	
	Fig.~\ref{fig:2x2cod} details the quality of fit in terms of how well the predicted accuracies relate to the actual accuracies. In each figure, values closer to the red diagonal line, represent individuals that have a better quality of fit. We can see that the higher the accuracy the closer our predicted values are to the red line. This is a by-product of the surrogate model management strategy we employ. In essence, we are less concerned with the accuracies of poorer-performing models, instead, we have taken a more greedy approach favouring models that are better performing. 
	
	Figure~\ref{fig:cod40} informs us why the $\times$40 magnification slightly under-performed in terms of the metrics as shown in Table~\ref{tab:surrogate_effectiveness}. We can see that for the predicted values, on the y-axis, there are quite several individuals predicted to have 80\% accuracy when the actual accuracies, for these individuals range from 30-80\% in terms of actual accuracy, as shown on the x-axis. A deeper dive revealed that this was a result of one particularly bad-performing run and was not present in the other 7 runs.


	\subsection{Analysis of the Energy Consumption}
	
	We perform a comparison of the estimated energy consumption of the \Surrogate{} and the \Expensive{} models using the Green-algorithm by Lannelongue et al.~\cite{lannelongue2021green}. Eq.~\ref{eqn::energy} is used to get this estimate,
	
	\begin{equation}
	\begin{aligned}
	E = r_t* (P_c * U + P_m) * PUE * PSF
	\end{aligned}
	\label{eqn::energy}
	\end{equation}
	
	\noindent where $E$ is the energy consumed in KW/h, $r_t$ is the runtime, $P_c$ is the power draw for the computing cores and depends on the CPU/GPU model, $U$ is the usage factor, $P_m$ is the power draw for the memory. $PUE$ is the Power Usage Efficiency and is a measure of how much additional energy is needed to operate the data centre. $PSF$ is the Pragmatic Scaling Factor and can be used to estimate the performance when multiple runs are taken into consideration. A conservative $PUE$ value of 1.67 was used, which represents the worldwide average power usage $PUE$ across all data centres~\cite{lannelongue2021green}. For simplicity, $PSF$ was kept as 1 as we are interested in the average runtime rather than the cumulative runtime.
	
	The real usage factor was minimal for both the \Surrogate{} and the \Expensive{} models, even during training for the \Surrogate{} model and additionally, the CPU energy consumption is negligible compared to that of the GPU, so we have reported just the energy saved by the GPU, overall 2.79 KW/h are saved using the \Surrogate{} model. Considering the total energy saved across all 4 magnifications for 8 runs each, we save approximately 89.28 KW/h: 25\% less energy is consumed using the \Surrogate{} approach.
	
	
	
	
	
	\subsection{Analysis of Genotype}

\begin{figure*}[htbp]
		\begin{subfigure}{0.42\linewidth}
			\includegraphics[width=\linewidth]{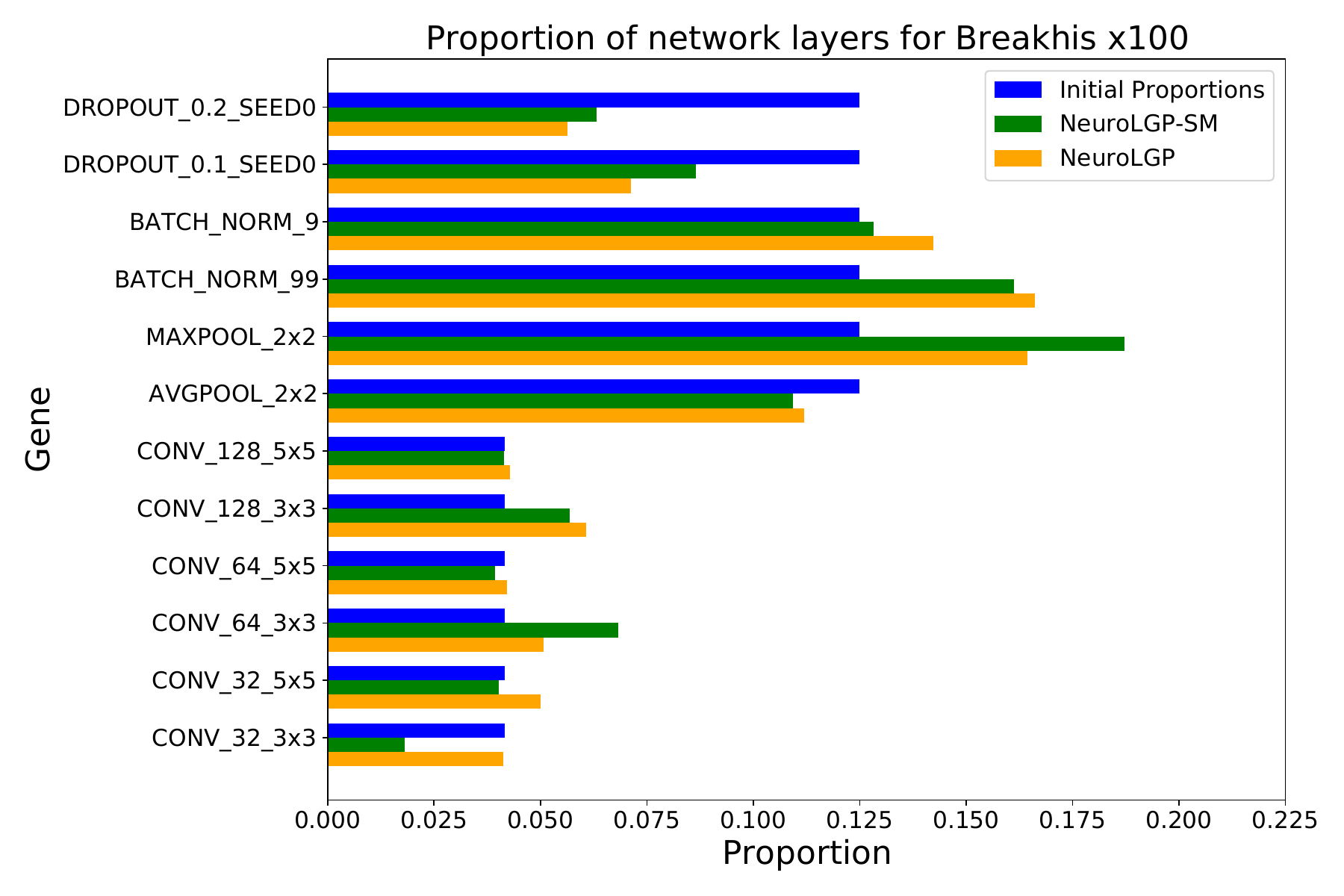}
			\caption{Breakhis $\times$100 dataset}
			\label{fig:geno100}
		\end{subfigure}
		\hfill
		\begin{subfigure}{0.42\linewidth}
			\includegraphics[width=\linewidth]{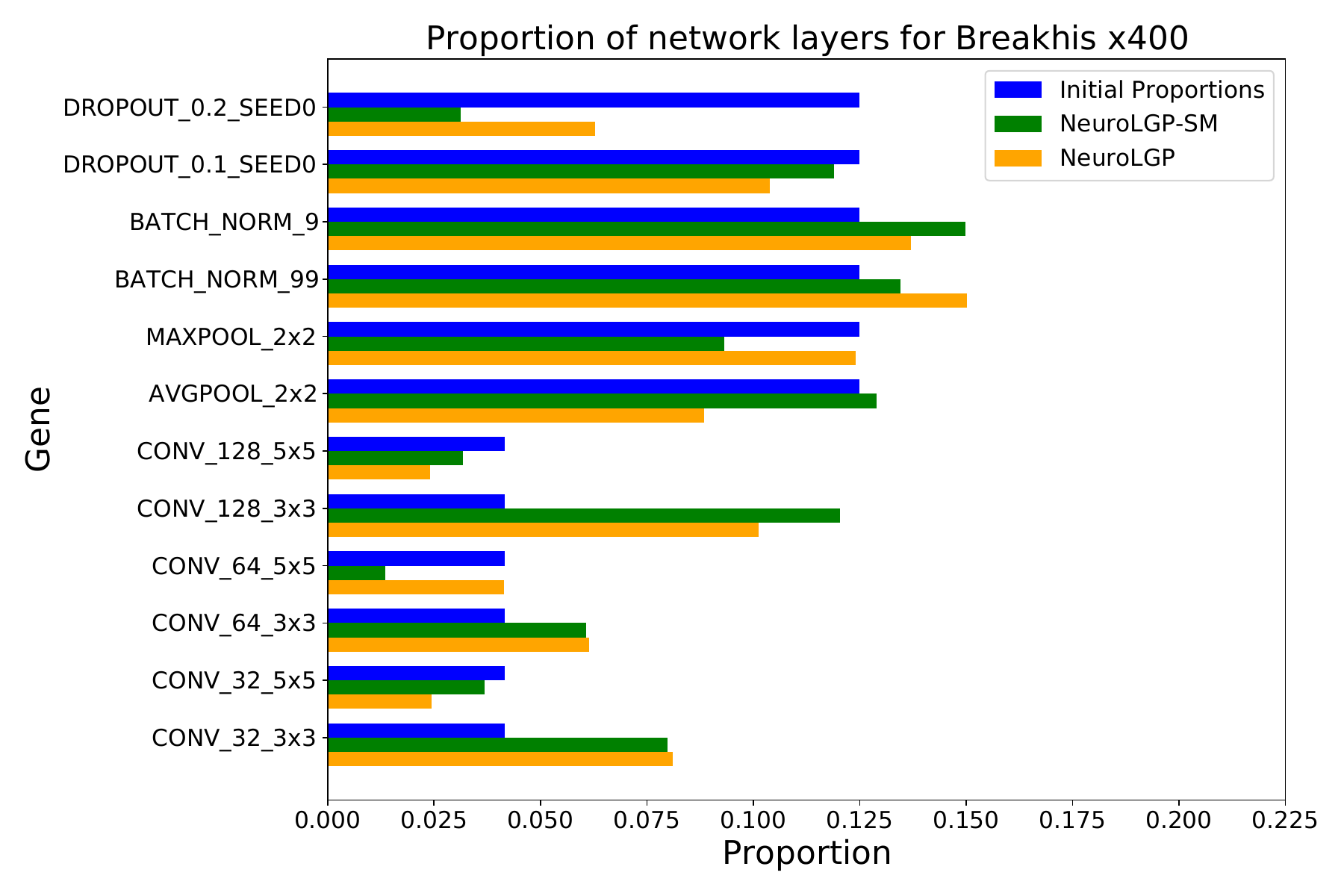}
			\caption{Breakhis $\times$400 dataset}
			\label{fig:geno400}
		\end{subfigure}
		
		\caption{Proportion of evolved network layers for initialisation (blue) and final generations for NueroLGP-SM (green) and NeuroLGP (orange) for $\times$100, and $\times$400 datasets}
		
		\label{fig:2x2geno}
	\end{figure*}

	Figs.~\ref{fig:2x2geno} (a) and (b) show the proportions of various genes for two of the four magnifications $\times$100 and $\times$400, respectively. Due to page constraints, we have shown only these two datasets, but our analysis for the same for the other two, unless stated otherwise in the text. Initial proportions (blue) represent the proportions of specific genes at initialisation (first generation). The \Surrogate{} (green) and \Expensive{} (orange) model are also represented and denote the proportions in the final generation across all 8 runs.
	The comparison we make will aim to show how the proportions the \Surrogate{} and \Expensive{} for specific layers change from initialisation (i.e., we compare green and orange proportions against blue).

	For the initial proportions, genes were proportioned based on their functional grouping. The four groups are dropout, batch normalisation, pooling, and convolutional layers. As such, each group makes up a quarter of the initial population and are subsequently divided again by specific genes. For instance, as there are 6 convolutional layer genes we divide $\frac{0.25}{6}$ to get the proportion for each convolutional layer. 
	
	Turning our attention to the \Surrogate{} and \Expensive{} model, we can see general trends in how the various gene groupings change by the final generation shown in green and orange for the NeuroLGP-SM and NeuroLGP, respectively. For instance, the dropout layer proportions sizes tend to decrease significantly by the final generation in the \Surrogate{} and \Expensive{} models. Similarly, the proportion of batch normalisation layers tends to increase. The pooling layers are more specific to the dataset. For instance, for $\times$100 there is an increase specifically for the max pooling layer (also found in $\times$200 but not shown). On the other hand, for $\times$400 there is a general decrease, albeit the two models differ on the specific pooling layer they decrease. Again, for the convolutional layers there is a more specific trend for particular datasets. While there are some slight increases and decreases in proportion size for $\times$40, $\times$100 and $\times$200, for $\times$400 there are notable increases in proportion sizes for $\times$400 for convolutional layers using a 3$\times$3 filter (Figure~\ref{fig:geno400}, CONV\_32\_3x3, CONV\_64\_3x3 and CONV\_128\_3x3).
	
	Some of these trends are unsurprising, we would expect that the proportion of dropout layers would be less, for instance, as CNN architectures are not as prone to overfitting~\cite{park2017analysis}. It would seem that the decrease in pooling layers in the $\times$400 is being compensated by an increase in convolutional layers, in other words, some of the dimension reduction is being handled by the convolutional layers rather than the pooling. Further research could investigate how these proportions change relative to different resolution sizes. Interestingly, while in many cases the change in proportions are consistent for both the \Surrogate{} and \Expensive{}, in some instances the proportion sizes differ significantly, for instance, CONV\_64\_5x5 as seen in  $\times$400. 
	This is likely a result of solutions converging at different local optima, however, further analysis of the fitness landscape would be beneficial in gaining more insight.
	
	\subsection{Limitations of our analysis}
	
	While a single run is the norm in terms of neuroevolution, we endeavoured to perform as many runs as possible, to add better confidence to our analysis. While we ran individual runs in parallel, both the \Surrogate{} and \Expensive{} approaches, for 8 runs each and for the 4 datasets, the total runtime amassed $\sim$50 GPU days. To conduct a full statistical analysis we would require another +50 GPU days of experimentation. Future work, could look into using different problem domains and different Deep learning models.
	
	
	

	\section{Conclusion}
	\label{sec::conclusion}
	
	
	In this work, we demonstrate a novel surrogate-assisted neurovolutionary approach, named Neuro-Linear Genetic Programming LGP surrogate model (NeuroLGP-SM). This approach makes use of Kriging Partial Least Squares (KPLS) to estimate the fitness of partially trained Deep Neural Networks (DNNs) using phenotypic distance vectors in a high-dimensional context. This scalable approach to surrogate-assisted neuroevolution is adept at finding robust and efficient architectures.
	We demonstrate that our approach is competitive or superior to other state-of-the-art handcrafted networks. Furthermore, we demonstrate that the proposed \Surrogate{} approach is 25\% more efficient in energy consumption, compared to a \Expensive{} variant that does not use surrogacy. Additionally, due to the unique encoding properties of our NeuroLGP approach, we can easily analyse the internal structures of the architectures, giving greater insight into the favorable components of the discovered architectures. 
	
	\section*{Acknowledgements}
	
	This publication has emanated from research conducted with the financial support of Science Foundation Ireland under Grant number 18/CRT/6049. The author has applied a CC BY public copyright licence to any Author Accepted Manuscript version arising from this submission.

	\bibliographystyle{abbrv}
	\bibliography{ref.bib}

\end{document}